\documentstyle[twoside,palatino,cl,epsfig,rotate]{article}

\let\cbstart\relax
\let\cbend\relax

\issue{27}{1}{2001}

\begin{document}

\bibliographystyle{fullname}

\title{Integrating Prosodic and Lexical Cues for Automatic Topic Segmentation}

\runningauthor{T\"ur, Hakkani-T\"ur, Stolcke, and Shriberg}
\runningtitle{Integrating Prosody for Topic Segmentation}

\author{G\"okhan T\"ur%
	 \thanks{Department of Computer Engineering, Bilkent University,
		Ankara, 06533, Turkey.
		E-mail: \{tur,hakkani\}@cs.bilkent.edu.tr.
		The research reported here was carried out while
		the authors were International Fellows at SRI International.}
       					& Dilek Hakkani-T\"ur\footnotemark[1] \\
	\affil{Bilkent University}	& \affil{Bilkent University} \\
		\\
	Andreas Stolcke%
         \thanks{Speech Technology and Research Laboratory, SRI International,
		333 Ravenswood Ave., Menlo Park, CA 94025, U.S.A.
	        E-mail: \{stolcke,ees\}@speech.sri.com.}
					& Elizabeth Shriberg\footnotemark[2] \\
	\affil{SRI International}	& \affil{SRI International} \\
}

\maketitle 

%\setlength{\baselineskip}{2\baselineskip}      % double-spacing

%\input{paper-body}
% changes in copy-editing:
% - spelling 
% - table formatting (rulers added)
% - citations inside parens use brackets

\renewcommand{\textfraction}{0.1}
\renewcommand{\floatpagefraction}{0.9}

\newcommand{\WT}{W_{\mbox{\scriptsize t}}}
\newcommand{\PLM}{P_{\mbox{\scriptsize LM}}}
\newcommand{\PDT}{P_{\mbox{\scriptsize DT}}}
\newcommand{\PHMM}{P_{\mbox{\scriptsize HMM}}}
\newcommand{\argmax}{\mathop{\rm argmax}}
\newcommand{\B}{B}	% boundary type variable
\newcommand{\T}{T}	% topic cluster variable

\newcommand{\0}{\kern 0.5em}	% digit-width space used in table alignment

\begin{abstract}
  We present a probabilistic model that uses both prosodic and lexical
  cues for the automatic segmentation of speech into topically
  coherent units. We propose two methods for combining lexical
  and prosodic information using hidden Markov models and decision
  trees. Lexical information is obtained from a speech recognizer, and
  prosodic features are extracted automatically from speech waveforms.
  We evaluate our approach on the Broadcast News corpus, using
  the DARPA-TDT evaluation metrics. Results show that the prosodic model
  alone is competitive with word-based segmentation methods. Furthermore,
  we achieve a significant reduction in error by combining the
  prosodic and word-based knowledge sources.
\end{abstract}

\section{Introduction}
{\bf Topic segmentation} is the task of automatically dividing a
stream of text or speech into topically homogeneous blocks.
That is, given a sequence of (written or spoken) words, the aim of
topic segmentation is to find the boundaries where topics change.
Figure~\ref{fig:example} gives an example of a topic change boundary
from a broadcast news transcript. Topic segmentation is an important task
for various
language understanding applications, such as information extraction
and retrieval, and text summarization.  In this paper, we present our
work on automatic detection of topic boundaries from speech input
using both prosodic and lexical information.

Other automatic topic segmentation systems have focused on written text and
have depended largely on lexical information. This approach is problematic
when segmenting speech.  Firstly, relying on word identities can propagate
automatic speech recognizer errors to the topic segmenter.
Secondly, speech lacks typographic cues, as shown in Figure~\ref{fig:example}:
there are no headers, paragraphs, sentence punctuation marks,
or capitalized letters.  Speech itself, on the other hand,
provides an additional, nonlexical knowledge source through its durational,
intonational, and energy characteristics, i.e., its {\bf prosody}.

Prosodic cues are known to be relevant to discourse structure in
spontaneous speech (cf.\ Section~\ref{sec:previous-prosody})
and can therefore be
expected to play a role in indicating topic transitions.  Furthermore,
prosodic cues by their nature are relatively unaffected by word
identity, and should therefore improve the robustness of lexical topic
segmentation methods based on automatic speech recognition.

\begin{figure}[t]
\fbox{\parbox{0.9\textwidth}{\small
\setlength{\baselineskip}{\normalbaselineskip}
\ldots tens of thousands of people are homeless in northern china tonight
after a powerful earthquake hit an earthquake registering six point
two on the richter scale at least forty seven people are dead few
pictures available from the region but we do know temperatures there
will be very cold tonight minus seven degrees {\bf $<$TOPIC\_CHANGE$>$}
peace talks expected to resume on monday in belfast northern ireland
former u. s. senator george mitchell is representing u. s. interests
in the talks but it is another american center senator rather who was
the focus of attention in northern ireland today here's a. b. c.'s
richard gizbert the senator from america's best known irish catholic
family is in northern ireland today to talk about peace and
reconciliation a peace process does not mean asking unionists or
nationalists to change or discard their identity or aspirations \ldots
}}
\vskip 0.5\baselineskip
\caption{An example of a topic boundary in a broadcast news word transcript.}
\label{fig:example}
\end{figure}

Topic segmentation
research based on prosodic information has generally relied on
hand-coded cues ({\citebrackets with the notable exception of
\cbstart
\namecite{HirschbergNakatani:98}}), or has not combined prosodic information
with lexical cues ({\citebrackets \namecite{Litman:95} is one example where
lexical information was combined with hand-coded prosodic features for
a related task}).
\cbend
Therefore, the present work
is the first that combines automatic extraction of both
lexical and prosodic information for topic segmentation.  

The general framework for combining lexical and prosodic cues for
tagging speech with various kinds of ``hidden'' structural information
is a further development of our earlier work on sentence segmentation
and disfluency detection for spontaneous speech
\cite{ShribergEtAl:eurospeech97,StoShr:icslp96,SOBS:icslp98},
conversational dialog tagging \cite{StolckeEtAl:CL2000},
and information extraction from broadcast news \cite{DilekEtAl:euro99}.

In the next section, we review previous work on topic
segmentation. In Section \ref{approach}, we describe our prosodic
and language models as well as methods for combining them. Section
\ref{results} reports our experimental procedures and results. 
We close with some general discussion (Section~\ref{discussion}) and 
conclusions (Section~\ref{conclusion}).

\section{Previous Work}
	\label{sec:previous}

Work on topic segmentation is generally based on two
broad classes of cues.
On the one hand, one can exploit the fact that topics are correlated 
with {\bf topical content-word usage},
and that global shifts in word usage are indicative of changes in topic.
Quite independently, {\bf discourse cues}, or linguistic devices
such as discourse markers, cue phrases, syntactic constructions,
and prosodic signals are employed by speakers (or writers) as generic
indicators of endings or beginnings of topical segments.
Interestingly, most previous work has explored either one or the other
type of cue, but only rarely both.
In automatic segmentation systems,
word usage cues are often captured by statistical language modeling and 
information retrieval techniques.
Discourse cues, on the other hand, are typically modeled with
rule-based approaches or classifiers derived by machine-learning
techniques (such as decision trees).

\subsection{Approaches Based on Word Usage}
	\label{sec:previous-topical}

Most automatic topic segmentation work based on text sources has
explored topical word usage cues in one form or other.
\namecite{Kozima:93} used mutual similarity of words in a sequence of
text as an indicator of text structure.
\namecite{Reynar:94} presented a method that finds topically similar
regions in the text by graphically modeling the distribution of word
repetitions.
The method of Hearst \shortcite{Hearst:94,Hearst:97}
uses cosine similarity in a word 
vector space as an indicator of topic similarity.

More recently, the U.S.~Defense Advanced Research Projects Agency (DARPA)
initiated the Topic Detection and Tracking (TDT)
program to further the state of the art in finding and following new
topics in a stream of broadcast news stories. One of the tasks
in the TDT effort is segmenting a news stream into individual stories.
Several of the participating systems rely essentially on word usage: 
\namecite{Yamron:98} model topics with unigram language models and their
sequential structure with hidden Markov models (HMMs).
\namecite{PonteCroft:97} extract related word sets for topic segments with
the information retrieval technique of local context analysis, and then
compare the expanded word sets.

\subsection{Approaches Based on Discourse and Combined Cues}
	\label{sec:previous-discourse}

Previous work on both text and speech has found that cue phrases or
discourse particles (items such as {\em now} or {\em by the way}), as well
as other lexical cues, can provide valuable indicators of structural
units in discourse \cite[among others]{Grosz:86,Passonneau:97}.

In the TDT framework,
the UMass ``HMM'' approach described in \namecite{Allan:98} uses an HMM
that models the initial, middle, and final sentences of a topic segment,
capitalizing on discourse cue words that indicate beginnings and ends of 
segments.  Aligning the HMM to the data amounts to segmenting it.

\namecite{Beeferman:99} combined a large set of automatically selected
lexical discourse cues in a maximum-entropy model.  They also incorporated
topical word usage into the model by building two statistical language
models: one static (topic independent) and one that adapts its
word predictions based on past words. They showed that the
log likelihood ratio of the two predictors behaves as an indicator of topic
boundaries, and can thus be used as an additional feature in the exponential
model classifier.

\cbstart
\subsection{Approaches Using Prosodic Cues}
	\label{sec:previous-prosody}

Prosodic cues form a subset of discourse cues in speech, reflecting systematic
duration, pitch, and energy patterns at topic changes and related
locations of interest.
A large literature in linguistics and related fields has shown
that topic boundaries (as well as similar entities such as paragraph boundaries
in read speech, or discourse-level boundaries in spontaneous speech)
are indicated prosodically in a manner that is similar to sentence or utterance
boundaries---only stronger.  Major shifts in topic typically show
longer pauses, an extra-high F0 onset or ``reset'', a higher maximum
accent peak, greater range in F0 and intensity \cite{%  
BrownEtAl:80,% 
GroszHirschberg:92,%
NakajimaAllen:93,%
GeluykensSwerts:93,%
Ayers:94,%
HirschbergNakatani:96,%
Nakajima:97,%
Swerts:97}
and shifts in speaking rate 
\cite{Brubaker:72,%
KoopmansDonzel:96,%
HirschbergNakatani:96}.
Such cues are known to be salient for human
listeners; in fact, subjects can perceive major discourse boundaries
even if the speech itself is made unintelligible via spectral
filtering \cite{SwertsEtAl:92}.

Work in automatic extraction and computational modeling of these
characteristics has been more limited, with most of the work in computational
prosody modeling dealing with  boundaries at the sentence level or
below. However, there have been some studies of discourse-level boundaries
in a computational framework.  They differ in various ways,
such as type of data (monolog or dialog, human-human or human-computer),
type of features (prosodic and lexical versus prosodic only), which features
are considered available (e.g., utterance boundaries or no boundaries),
to what extent features are automatically extractable and normalizable,
and the machine learning approach.
Because of these vast difference the overall results cannot be compared directly
to each other or to our work, but we describe three of the approaches
briefly here. 

An early study by \namecite{Litman:95} used hand-labeled prosodic boundaries
and lexical information, 
but applied machine learning to a training corpus and tested on unseen data. 
The researchers combined
pause, duration, and hand-coded intonational boundary information
with lexical information from cue phrases (such as {\em and} and {\em so}). 
Additional knowledge sources included complex relations,
such as coreference of noun phrases.
Work by \namecite{SweOst:97} used prosodic features
that in principle could be extracted automatically,
such as pitch range, to classify utterances
from human-computer task-oriented dialog into two categories:
initial or noninitial
in the discourse segment. The approach used CART-style decision trees to model
the prosodic features, as well as various lexical features that
could also in principle be estimated automatically. In this case, utterances
were presegmented, so the task was to classify segments rather than
find boundaries in continuous speech;
the features also included some (such as type of boundary
tone) that may not be easy to extract robustly across speaking styles. 
Finally, \namecite{HirschbergNakatani:98}
proposed a prosody-only front end for tasks such as audio browsing and
playback, which could segment continuous audio input into meaningful
information units.  They used automatically extracted pitch, energy,
and ``other'' features (such as the cross-correlation value used by
the pitch tracker in determining the estimate of F0) as inputs to
CART-style trees, and aimed to predict major discourse-level
boundaries. They found various effects of frame window length and
speakers, but concluded overall that prosodic cues could be useful for
audio browsing applications.

\cbend

\section{The Approach}
\label{approach}

Topic segmentation in the paradigm used in this study and others
\cite{Allan:98}
proceeds in two phases.  In the first phase, the input is divided into
contiguous strings of words assumed to belong to the same topic.  We
refer to this step as {\bf chopping}.  For example, in textual input,
the natural units for chopping are sentences (as can be inferred from
punctuation and capitalization), since we can assume that topics do not
change in mid-sentence.\footnote{Similarly, it is sometimes assumed for
topic-segmentation purposes that topics change only at paragraph
boundaries \cite{Hearst:97}.}
For continuous speech input, the
choice of chopping criteria is less obvious;
we compare several possibilities in our
experimental evaluation.  Here, for simplicity, we will use
{\bf sentence} to refer to units of chopping, regardless of the
criterion used.

In the second phase, the sentences are further
grouped into contiguous stretches belonging to one topic, i.e., the
sentence boundaries are classified into {\bf topic boundaries} and
{\bf nontopic boundaries}.\footnote{We do not consider the problem of
detecting recurring, discontinuous instances of the same topic, a task
known as {\bf topic tracking} in the TDT paradigm \cite{TDT2}.}
Topic segmentation is thus reduced to a boundary classification
problem.  We will use $\B$ to denote the string of binary boundary
classifications.  Furthermore, our two knowledge sources are the
(chopped) word sequence $W$ and the stream of prosodic features $F$.
Our approach aims to find the segmentation $\B$ with highest
probability given the information in $W$ and $F$

\begin{equation}
        \argmax_\B P(\B | W, F) \quad
\label{max}
\end{equation}

\noindent using statistical modeling techniques. 

In the following subsections, we first describe 
the prosodic model of the dependency
between prosody $F$ and topic segmentation $\B$; then, the language
model relating words $W$ and $\B$; and finally, two approaches for
combining the models.

\subsection{Prosodic Modeling}
\label{prosody}

The job of the prosodic model is to estimate the posterior probability
\cbstart
(or, alternatively, likelihood)
\cbend
of a topic change at a given word boundary, based on prosodic features 
extracted from the data.   For the prosodic model to be 
effective, one must devise suitable, automatically extractable 
\cbstart
features. Feature values extracted from a corpus can then be used
in training probability estimators and 
to select a parsimonious subset of features for modeling purposes.
\cbend
We discuss each of these steps in turn in the following sections.

\subsubsection{Features}
We started with a large collection of features capturing two major
aspects of speech prosody, similar to our previous work
\cite{ShribergEtAl:eurospeech97}:
\begin{itemize}
\item
        Duration features: duration of pauses, duration of final vowels
\cbstart
		and final rhymes,\footnote{The rhyme is the part of a 
		syllable that comprises the nuclear phone (typically a
			vowel) and any following phones. This is the part
			of the syllable
			most typically affected by lengthening.}
\cbend
		and versions of these features normalized both
                for phone durations and speaker statistics.
\item
        Pitch features: fundamental frequency (F0) patterns preceding
		and following the boundary, F0 patterns across the boundary,
                and pitch range relative to the speaker's baseline.
		We processed the 
		raw F0 estimates ({\citebrackets
		obtained with ESPS signal processing
		software from \namecite{ESPS}}), with robustness-enhancing
		techniques developed by \namecite{Sonmez:98}.
		
\end{itemize}
We did not use amplitude- or energy-based features since exploratory
work showed these to be much less reliable than duration and pitch
and largely redundant given the above features. One reason for
omitting energy features is that,
unlike duration and pitch, energy-related measurements vary with
channel characteristics.  Since channel properties vary widely in
broadcast news, features based on energy measures can correlate with
shows, speakers, and so forth, rather than with the structural locations
in which we were interested.

We included features that, based on the descriptive literature,
should reflect breaks in the temporal and intonational
contour. We developed versions of such features that could be defined
at each interword boundary, and that could be extracted by
completely automatic means (no human labeling).  Furthermore, the
features were designed to be as independent of
word identities as possible, for robustness to imperfect recognizer output.
A brief characterization of the informative features for the
segmentation task is given with our results in
Section~\ref{sec:dt-prosody-only}.
Since the focus here is on computational modeling we refer the reader to
a companion paper \cite{ShribergEtAl:specom2000} for a 
detailed description of the acoustic processing and prosodic feature
extraction.

\subsubsection{Decision trees}
Any of a number of probabilistic classifiers (such as neural networks,
exponential models, or na{\"\i}ve Bayes networks) could be used as posterior
probability estimators.  As in past prosodic modeling work
\cite{ShribergEtAl:eurospeech97}, we chose CART-style decision trees
\cite{Breiman:84}, as implemented by the IND package \cite{IND},
because of their ability to model feature interactions, 
to deal with missing features, and to handle large amounts of 
training data.
The foremost reason for our preference for decision trees, however,
is that the learned models can be inspected and diagnosed
by human investigators. This ability is crucial for understanding what and
how features are used, and for debugging the feature extraction
\cbstart
process itself.\footnote{Interpreting large trees can be a 
daunting task.  However, the decision questions near the tree root
are usually interpretable, or, when nonsensical, usually indicate problems
with the data.
Furthermore, as explained in Section~\ref{sec:dt-prosody-only},
we have developed simple statistics that give an overview of feature usage
throughout the tree.}
\cbend

Let $F_i$ be the features extracted from a window around the $i$th
potential topic boundary (chopping boundary), and let $\B_i$ be the
boundary type (boundary/no-boundary) at that position.  We trained
decision trees to predict the $i$th boundary type, i.e.,
to estimate $P(\B_i | F_i, W)$.  
The decision is only weakly conditioned
on the word sequence $W$, insofar as some of the prosodic features
depend on the phonetic alignment of the word models (which we will denote
with $\WT$).  We can
thus expect the prosodic model estimates to be robust to recognition
errors. The decision tree paradigm also allows us to add, and
automatically select, other (nonprosodic) features that might be
relevant to the task.

\subsubsection{Feature selection}
The greedy nature of the decision tree learning algorithm implies that
larger initial feature sets can give worse results than smaller
subsets.  Furthermore, it is desirable to remove redundant features
for computational efficiency and to simplify the interpretation of
results.  For this purpose we developed an iterative feature selection
\cbstart
``wrapper'' algorithm \cite{JohnEtAl:94}
\cbend
that finds useful, task-specific feature subsets.  The algorithm
combines elements of a brute-force search with previously determined
heuristics about good groupings of features.
The algorithm proceeds in two phases: In the first phase, the number of
features is reduced by leaving out one feature at a time during tree
construction.  A feature whose removal increases performance
is marked as to be avoided.
The second phase then starts with the reduced feature set 
and performs a beam search over all possible subsets to maximize
tree performance.

We used entropy
reduction in the overall tree (after cross-validation pruning) as a metric for
comparing alternative feature subsets. Entropy reduction is the difference
in entropy between the prior class distribution and the
\cbstart
posterior distribution estimated by the tree, as measured on a held-out
\cbend
set; it is a more
fine-grained metric than classification accuracy, and is also more
relevant to the model combination approach described later.

\subsubsection{Training data}
To train the prosodic model, we automatically aligned and extracted
features from 70~hours (about 700,000~words) of the Linguistic Data Consortium
(LDC) 1997 Broadcast News (BN) corpus.  Topic boundary information
determined by human labelers was extracted from the SGML markup
that accompanies the word transcripts of this corpus. The word transcripts
were aligned automatically with the acoustic waveforms to obtain pause
and duration information, using the SRI Broadcast News
recognizer \cite{Sankar:darpa98}.

\subsection{Lexical Modeling}
\label{language}

Lexical information in our topic segmenter is captured by 
statistical language models (LMs) embedded in an HMM.
The approach is an extension of the topic segmenter
developed by Dragon Systems for the TDT2 effort \cite{Yamron:98}, 
which was based purely on topical word distributions.
We extend it to also capture lexical and (as described in
Section~\ref{combined}) prosodic discourse cues.

\subsubsection{Model structure}

\begin{figure}[t]
\psfig{figure=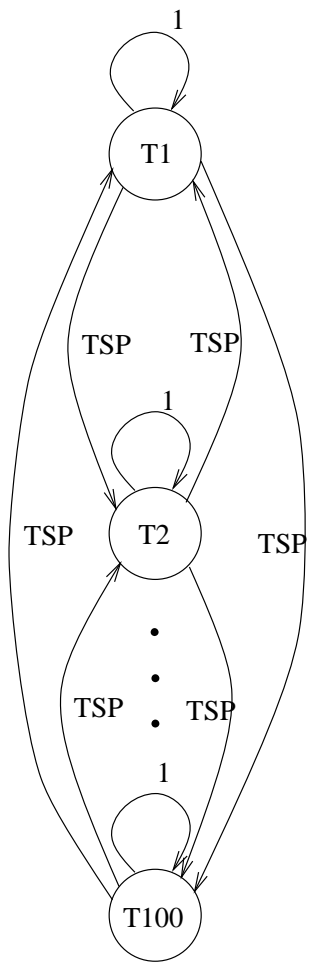}
\vskip 0.5\baselineskip
\caption{Structure of the basic HMM developed by Dragon for the TDT Pilot
Project. The labels on the arrows indicate the
transition probabilities. TSP represents the topic switch
penalty.}
\label{fig:dragonhmm}
\end{figure}

The overall structure of the model is that of an HMM \cite{Rabiner:86}
in which the states correspond to topic clusters $\T_j$,
and the observations are sentences (or chopped units) $W_1, \ldots, W_N$.
The resulting HMM, depicted in Figure \ref{fig:dragonhmm}, forms a
complete graph, allowing for transitions between any two topic clusters.
Note that it is not necessary that the topic clusters correspond exactly
to the actual topics to be located; for segmentation purposes it is
sufficient that two adjacent actual topics are unlikely to be mapped to the
same induced cluster.
The observation likelihoods for the HMM states, $P(W_i|\T_j)$,
represent the probability of generating a given sentence $W_i$ in a
particular topic cluster $\T_j$.

We automatically constructed 100~topic cluster LMs, using the
multipass $k$-means algorithm described in \namecite{Yamron:98}.
Since the HMM emissions are meant to model the topical usage of words,
but not topic-specific syntactic structures, the LMs consist of unigram
distributions that exclude stop words
(high-frequency function and closed-class words).
To account for unobserved words we interpolate the topic cluster-specific
LMs with the global unigram LM obtained from the entire training data.
The observation likelihoods of the HMM states are then computed from these
smoothed unigram LMs.

All HMM transitions within the same topic cluster are given probability one,
whereas all transitions between topics are set to a global {\bf topic switch
penalty} (TSP) that is optimized on held-out training data. The TSP
parameter allows trading off between false alarms and misses.
Once the HMM is trained, we use the Viterbi algorithm
\cite{Viterbi:67,Rabiner:86}
to search for the best state sequence and corresponding segmentation.
Note that the transition probabilities in the model are not normalized
to sum to one; this is convenient and permissible since the output of the
Viterbi algorithm depends only on the relative weight of the transition 
weights.

\begin{figure}[t]
\psfig{figure=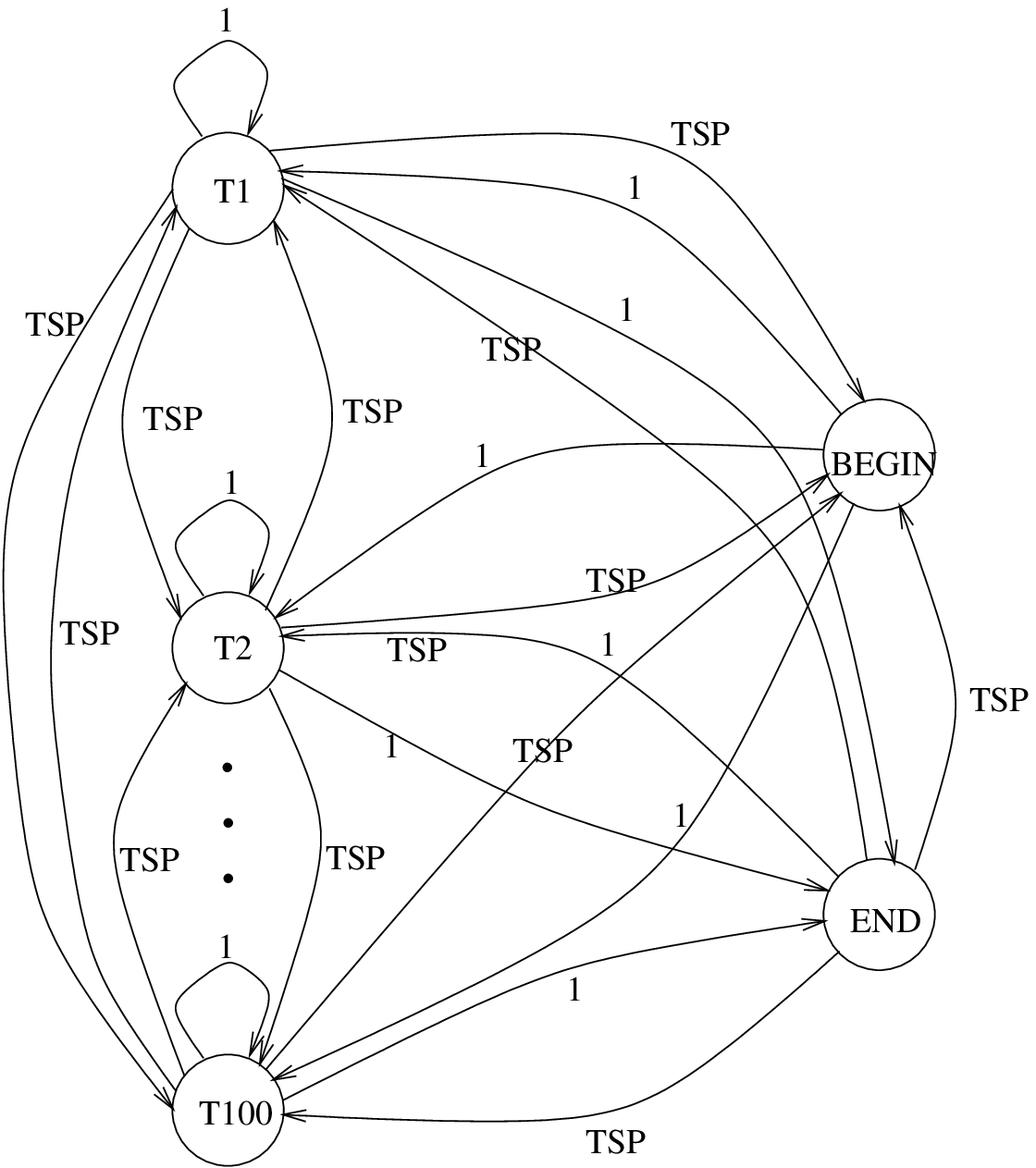}
\vskip 0.5\baselineskip
\caption{Structure of an HMM with topic BEGIN and END states.
TSP represents the topic switch penalty.}
\label{fig:hmm}
\end{figure}

We augmented the Dragon segmenter with additional states and transitions to
also capture lexical discourse cues.  In particular, we wanted to
model the initial and final sentences in each topic segment, as these
often contain formulaic phrases and keywords used by broadcast speakers
({\em From Washington, this is \ldots}, {\em And now \ldots}).
We added two additional states {\em BEGIN} and {\em END} to the HMM
(Figure~\ref{fig:hmm}) to model these sentences.
Likelihoods for the BEGIN and END states are
obtained as the unigram language model probabilities of the
initial and final sentences, respectively, of the topic segments
in the training data.
Note that a single BEGIN and END state are shared for all topics.
Best results were obtained by making traversal of these
states optional in the HMM
topology, presumably because some initial and final sentences are better
modeled by the topic-specific LMs.

The resulting model thus effectively combines the Dragon and UMass HMM
topic segmentation approaches described in \namecite{Allan:98}.
In preliminary experiments, 
we observed a 5\% relative reduction in segmentation error
with initial and final states
over the baseline HMM topology of Figure~\ref{fig:dragonhmm}.
Therefore, all results reported later use an HMM topology with initial and
final states.
\cbstart
Note that, since the topic-initial and topic-final states are optional,
our training of the model is suboptimal. Instead of labeling all
topic-initial and topic-final training sentences as data for the
corresponding state, we would expect further improvements by training the
HMM in unsupervised fashion using the Baum-Welch algorithm
\cite{Baum:70,Rabiner:86}.
\cbend

\subsubsection{Training data}
Topic unigram language models were trained from the pooled TDT Pilot
and TDT2 training data \cite{TDT2:darpa99}, covering transcriptions of
broadcast news from January 1992 through June 1994 and from January
1998 through February 1998, respectively. These corpora are similar in
style, but do not overlap with the 1997 LDC BN corpus from which
we selected our prosodic training data and the evaluaton test set.
For training the language models we removed stories with fewer than 300 and
more than 3000 words, leaving 19,916 stories with an average length of 538
words (including stop words).

\subsection{Model Combination}
\label{combined}

We are now in a position to describe how lexical and prosodic information
can be combined for topic segmentation.
As discussed before, the LMs in the HMM capture topical word usage as 
well as lexical discourse cues at topic transitions, whereas 
a decision tree models prosodic discourse cues.
We expect that these knowledge sources are largely independent, so 
their combination should yield significantly improved performance.

\begin{figure}[t]
\psfig{figure=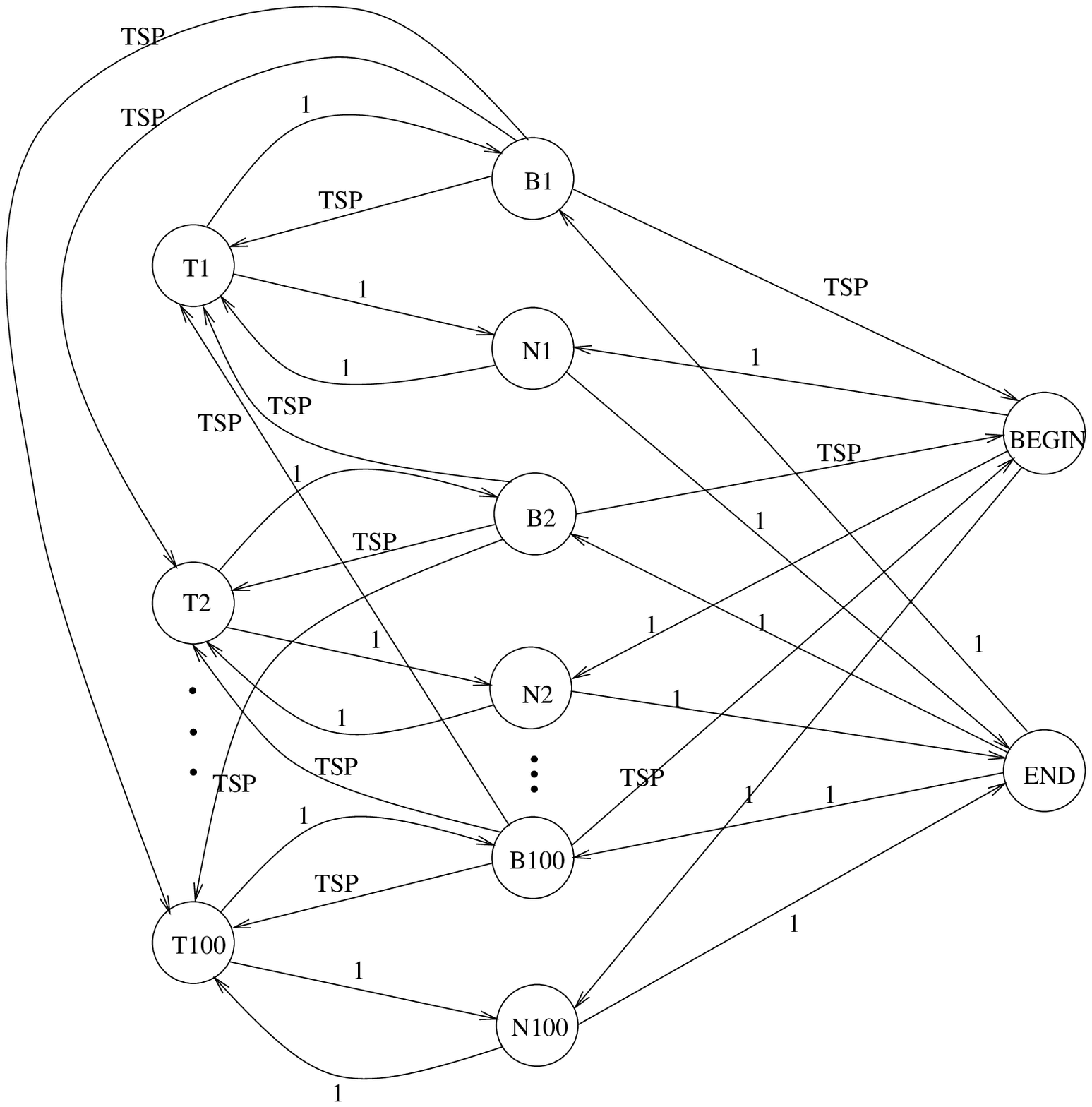,width=\textwidth}
\vskip 0.5\baselineskip
\caption{Structure of the final HMM with fictitious boundary states
used for combining language and prosodic models. In the figure,
states B1, B2, \ldots, B100 represent the presence of a topic boundary,
whereas states N1, N2, \ldots, N100 represent topic-internal sentence
boundaries.  TSP is the topic switch penalty.}
\label{fig:finalhmm}
\end{figure}

Below we present two approaches for building
a combined statistical model that performs topic segmentation
using all available knowledge sources.
For both approaches it is convenient to
associate a ``boundary'' pseudotoken with each potential topic boundary
(i.e., with each sentence boundary). Correspondingly, we introduce,
into the HMM, new states that emit these boundary tokens.
No other states emit boundary tokens; therefore each sentence boundary
must align with one of the boundary states in the HMM.
As shown in Figure~\ref{fig:finalhmm}, there are two boundary states for
each topic cluster, one representing a topic transition and the other
representing a topic-internal transition between sentences.
Unless otherwise noted, the observation likelihoods for the boundary 
states are set to unity.

\newcommand{\Bstate}{{\bf B}}
\newcommand{\Nstate}{{\bf N}}

The addition of boundary states allows us to compute the model's
prediction of topic changes as follows:
Let $\Bstate_1, \ldots, \Bstate_C$ denote the topic boundary states
and, similarly, let $\Nstate_1, \ldots, \Nstate_C$ denote
the nontopic boundary states, where $C$ is the number of topic clusters.
Using the forward-backward algorithm for HMMs \cite{Rabiner:86},
we can compute $P(q_i = \Bstate_j | W)$ and $P(q_i = \Nstate_j | W)$, the 
posterior probabilities that one of these states is occupied at boundary $i$.
The model's prediction of a topic boundary is simply the sum over the 
corresponding state posteriors:
\begin{eqnarray}
\PHMM(B_i = {\bf yes} | W ) & = & \sum_{j = 1}^{C} P(q_i = \Bstate_j | W) 
							\label{eq:yes} \\
\PHMM(B_i = {\bf no} | W ) & = & \sum_{j = 1}^{C} P(q_i = \Nstate_j | W) 
							\label{eq:no} \\
			& = & 1 - \PHMM(B_i = {\bf yes} | W ) \nonumber 
\end{eqnarray}

\subsubsection{Model combination in the decision tree} 
\label{combined-dt}
Decision trees allow the training of a single classifier that takes both
lexical and prosodic features as input, provided we can compactly encode
the lexical information for the decision tree.
We compute the posterior probability  $\PHMM(\B_i = {\bf yes} | W)$
as shown above, to summarize the HMM's belief in a topic
boundary based on all available lexical information $W$.
The posterior value is then used as an additional input feature to the
prosodic decision tree, which is trained in the usual manner.
During testing, we declare a topic boundary whenever the tree's overall
posterior estimate $\PDT(\B_i|F_i, W)$ exceeds some threshold.
The threshold may be varied to trade off false alarms for miss errors,
or to optimize an overall cost function.

Using HMM posteriors as decision tree features is similar in spirit to
the knowledge source combination approaches used by \namecite{Beeferman:99}
and \namecite{Reynar:99}, who also used the output of a topical word usage
model as input to an overall classifier.
In previous work \cite{SOBS:icslp98} we used the present approach as one
of the knowledge source combination strategies for sentence and disfluency
detection in spontaneous speech.

\subsubsection{Model combination in the HMM}
\label{combined-hmm}

An alternative approach to knowledge source combination uses the HMM
as the top-level model.
In this approach, the prosodic decision tree is used to estimate
likelihoods for the boundary states of the HMM,
thus integrating the prosodic evidence into the HMM's segmentation
decisions.

\newcommand{\Tstate}{{\bf T}}

More formally, let $Q = (r_1, q_1, \ldots, r_i, q_i, \ldots, r_N, q_N)$ be 
a state sequence through the HMM.  The model is constructed such that 
the states $r_i$ representing topic (or BEGIN/END) clusters alternate 
with the states $q_i$ representing boundary decisions.
As in the baseline model, the likelihoods of the topic cluster states
$\Tstate_j$ account for the lexical observations:
\begin{equation}
	P(W_i | r_i = \Tstate_j) = P(W_i | \T_j)
\end{equation}
as estimated by the unigram LMs.
Now, in addition, we let the likelihood of the boundary state at position
$i$ reflect the prosodic observation $F_i$.
\cbstart
Recall that, like $W_i$, $F_i$ refers to complete sentence units;
specifically, $F_i$ denotes the prosodic features of the $i$th boundary
between such units.
\cbend
\begin{equation}
\left .
\begin{array}{rcl}
	P(F_i | q_i = \Bstate_j, W) & = & P(F_i | {\B_i = {\bf yes}}, W) \\
	P(F_i | q_i = \Nstate_j, W) & = & P(F_i | {\B_i = {\bf no}}, W)
\end{array}
\right \} \mbox{for all $j = 1, \ldots, C$}
\end{equation}
Using this construction, the product of all state likelihoods will
give the overall likelihood, accounting for both lexical and prosodic 
observations:
\begin{equation}
	\prod_{i = 1}^{N} P(W_i | r_i) \prod_{i = 1}^{N} P(F_i | q_i, W)
		= P(W, F | Q )
\end{equation}
Applying the Viterbi algorithm to the HMM will thus return
the most likely segmentation conditioned on both words and prosody,
which is our goal.

Although decomposing the likelihoods as shown allows prosodic observations
to be conditioned on the words $W$, we use only the phonetic alignment
information $W_t$ from the word sequence $W$ in our prosodic models,
ignoring the word identities,
so as to make them more robust to recognition errors.

The likelihoods $P(F_i|\B_i,\WT)$ for the boundary states can
now be obtained from the prosodic decision tree.
Note that the decision tree estimates posteriors $\PDT(\B_i|F_i,\WT)$.
These can be converted to likelihoods using Bayes rule as in
\begin{equation}
P(F_i | \B_i,\WT) = { P(F_i|\WT) \PDT(\B_i|F_i,\WT) \over P(\B_i|\WT) } \quad .
\end{equation}
The term $ P(F_i|\WT)$ is a constant for all decisions $B_i$ and can thus
be ignored when applying the Viterbi algorithm.
\cbstart
Next, we approximate $P(\B_i|\WT) \approx P(\B_i)$, justified by the fact
that the $\WT$ contains only information about start and end times of phones and
words, but not directly about word identities.
\cbend
Instead of explicitly dividing the posteriors we 
prefer to downsample the training set to make
$P(\B_i = {\bf yes}) = P(\B_i = {\bf no}) = {1 \over 2}$.
A beneficial side effect of this approach is that
the decision tree models the lower-frequency events (topic boundaries)
in greater detail than if presented with the raw, highly skewed class 
distribution.

As is often the case when combining probabilistic models of different 
types, it is advantageous to weight the contributions of the 
language models and the prosodic trees relative to each other.
We do so by introducing a tunable {\bf model combination weight} (MCW),
and by using $\PDT(F_i|\B_i, \WT)^{\rm MCW}$ as the effective prosodic
likelihoods.  The value of MCW is optimized on held-out data.

\section{Experiments and Results}
\label{results}

To evaluate our topic segmentation models we carried out experiments in
the TDT paradigm.  We first describe our test data and the evaluation
metrics used to compare model performance.
We then give results obtained with individual knowledge sources,
followed by results using the combined models.

\subsection{Test Data}
We evaluated our system on three hours (6 shows, about 53,000 words)
of the 1997 LDC BN corpus.
The threshold for the model combination in the decision tree and the
topic switch penalty were optimized on the larger development training set
of 104~shows, which includes the prosodic model
training data. The MCW for the model combination in the HMM was
optimized using a smaller held-out set of 10 shows of about
85,000 words total size, separate from the prosodic model training
data.

We used two test conditions: forced alignments using the true words,
and recognized words as obtained by a simplified version of the SRI
Broadcast News recognizer \cite{Sankar:darpa98}, with a word error
% /home/spon3/sobs/Broadcast/nbest/4gram.9-0.sclite
rate of 30.5\%.

Our aim in these experiments was to use fully automatic recognition
and processing wherever possible.  For practical reasons, we 
departed from this strategy in two areas.  First, for word recognition, we 
used the acoustic waveform segmentations provided with the corpus
(which also included the location of nonnews material, such as commercials
and music).  Since current BN
recognition systems perform this segmentation automatically with very good
accuracy and with only a few percentage points penalty in word error rate
\cite{Sankar:darpa98}, we felt the added
complication in experimental setup and evaluation was not justified.

Second, for prosodic modeling, we used information from the corpus markup
concerning speaker changes and the identity of frequent speakers (e.g.,
news anchors).
Automatic speaker segmentation and labeling is
possible, though with errors \cite{NIST:euro99}.
Nevertheless, our use of speaker labels was motivated by the fact
that meaningful
prosodic features may require careful normalization by speaker, and
unreliable speaker information would have made the analysis
of prosodic feature usage much less meaningful.

\subsection{Evaluation Metrics}

We have adopted the evaluation paradigm used by the
TDT2---Topic Detection and Tracking Phase~2
\cite{TDT2} program, allowing fair comparisons of various approaches
both within this study and with respect to other recent work.  Segmentation 
accuracy was measured using TDT evaluation software from NIST, which
implements a variant of an evaluation metric suggested by 
\namecite{Beeferman:99}.

The TDT segmentation metric is different from those used in most previous
topic segmentation work, and therefore merits some discussion.
It is designed to work on data streams without any
potential topic boundaries, such as paragraph or sentence boundaries, being
given a priori. It also gives proper partial credit to segmentation
decisions that are close to actual boundaries; for example, placing a
boundary one word from an actual boundary is considered a lesser error than
if the hypothesized boundary is off by, say, 100 words.

The evaluation metric reflects the probability that two positions in
the corpus probed at random and separated by a distance of $k$ words 
are correctly classified as belonging to the same story or not.
If the two words belong to the same topic
segment, but are erroneously claimed to be in different topic segments by
the segmenter, then this will increase the system's {\bf false alarm}
probability. Conversely, if the two words are in different topic
segments, but are erroneously marked to be in the same segment, this
will contribute to the {\bf miss} probability.
The false alarm and miss rates are defined as averages over all
possible probe positions with distance $k$.

\cbstart
Formally, miss and false alarm rates are computed as\footnote{
The definitions are those from \namecite{TDT2}, but have 
been simplified and edited for clarity.}
\cbend
\begin{eqnarray}
P_{\it Miss} & = &
		\frac{\sum_{s} \sum_{i=1}^{N_s-k}d^s_{\it hyp}(i,i+k)
				\times (1-d^s_{\it ref}(i,i+k))}
		{\sum_{s}\sum_{i=1}^{N_s-k}(1-d^s_{\it ref}(i,i+k))} \\
P_{\it FalseAlarm} & = & 
		\frac{\sum_{s} \sum_{i=1}^{N_s-k}(1-d^s_{\it hyp}(i,i+k))
				\times d^s_{\it ref}(i,i+k)}
		{\sum_{s}\sum_{i=1}^{N_s-k}d^s_{\it ref}(i,i+k)}
\end{eqnarray}
where the summation is over all broadcast shows $s$
and word positions $i$ in the test corpus and where
\[
	d^s_{\it sys}(i,j) = \left \{ \begin{array}{ll}1 &
			\mbox{\parbox[t]{3in}{
			if words $i$ and $j$ in show $s$
			are deemed by {\it sys} to be within the same story}}\\
					0 & \mbox{otherwise}
				\end{array}\right.
\]
Here {\it sys} can be {\it ref} to denote the reference (correct)
segmentation, or {\it hyp} to denote the segmenter's decision.

For audio sources an analogous metric is defined where
segmentation decisions (same or different topic) are probed at
a time-based distance $\Delta$:
\newcommand{\D}{{\rm d}}
\begin{eqnarray}
P_{\it Miss} & = &
	\frac{\sum_{s} \int_{t=0}^{T_s-\Delta} d^s_{\it hyp}(t,t+\Delta)
				\times (1-d^s_{\it ref}(t,t+\Delta)) \D t}
       	   {\sum_{s} \int_{t=0}^{T_s-\Delta}(1-d^s_{\it ref}(t,t+\Delta))\D t}
	\\
P_{\it FalseAlarm} & = &
	\frac{\sum_{s} \int_{t=0}^{T_s-\Delta} (1-d^s_{\it hyp}(t,t+\Delta))
			\times d^s_{\it ref}(t,t+\Delta) \D t}
	     {\sum_{s} \int_{t=0}^{T_s-\Delta} d^s_{\it ref}(t,t+\Delta) \D t}
\end{eqnarray}
where the integration is over the entire duration of all 
stories of the shows in the test corpus, and where
\[
	d^s_{\it sys}(t_1,t_2) = \left \{ \begin{array}{ll} 1 &
			\mbox{\parbox[t]{3in}{if times $t_1$ and $t_2$
			in show $s$ are deemed by {\it sys} to be within
			the same story}}\\
		 0 & \mbox{otherwise}
		\end{array}\right.
\]

We used the same parameters as used in the
official TDT2 evaluation: $k = 50$ and $\Delta = 15 \mbox{ seconds}$.
Furthermore, again following NIST's evaluation procedure, we combine
miss and false alarm rates into a single {\bf segmentation cost} metric
\begin{equation}
C_{\it Seg} = C_{\it Miss} \times P_{\it Miss} \times P_{\it seg} +
	 C_{\it FalseAlarm} \times P_{\it FalseAlarm} \times (1-P_{\it seg})
\end{equation}
where the 
$C_{\it Miss} = 1 $ is the cost of a miss,
$C_{\it FalseAlarm} = 1$ is the cost of a false alarm,
and $P_{\it Seg} = 0.3$ is the a priori probability of a segment
being within an interval of $k$ words or $\Delta$ seconds on
the TDT2 training corpus.\footnote{
Another parameter in the NIST evaluation is the
deferral period, i.e., the amount of look-ahead before a
segmentation decision is made. In all our experiments we allowed unlimited 
deferral, effectively until the end of the news show being processed.}

\subsection{Chopping}
Unlike written text, the output of the automatic speech recognizer
contains no sentence boundaries.  Therefore, chopping text into
(pseudo)sentences is a nontrivial problem when processing
speech.  Some presegmentation into roughly sentence-length units is
necessary since otherwise the observations associated with HMM states
\cbstart
would comprise too few words to give robust likelihoods of topic choice,
\cbend
causing poor performance.

We investigated chopping criteria based on a fixed number of words (FIXED),
at speaker changes (TURN), at pauses
(PAUSE), and, for reference, at actual sentence boundaries (SENTENCE)
obtained from the transcripts.  Table~\ref{chopping} gives the error rates
for the four conditions, using
the true word transcripts of the larger development data set.  For the
PAUSE condition, we empirically determined an optimal minimum pause
duration threshold to use.  Specifically, we considered pauses
exceeding 0.575 of a second as potential topic boundaries in this (and all
later) experiments.  For the FIXED condition, a block length of 10~words
was found to work best.

\begin{table}[t]
\tcaption{Segmentation error rates for various chopping criteria,
using true words of the larger development data set.}
\label{chopping}
\begin{tabular}{lccc}
Chopping Criterion & $P_{\it Miss}$ & $P_{\it FalseAlarm}$ & $C_{\it Seg}$ \\
\hline
FIXED &0.5688&0.0639&0.2153\\
TURN &0.6737&0.0436&0.2326\\
SENTENCE &0.5469&0.0557&0.2030\\
PAUSE &0.5111&0.0688&0.2002 \\  
\end{tabular}
\end{table}

We conclude that a simple prosodic feature, pause duration, is an
excellent criterion for the chopping step, giving comparable or better
performance than standard sentence boundaries. Therefore, we used pause
duration as the chopping criterion in all further experiments.

\subsection{Source-specific Model Tuning}

As mentioned earlier, the segmentation models contain global
parameters
(the topic transition penalty of the HMM and the posterior threshold for
the combined decision tree)
to trade false alarms for miss errors.
Optimal settings for these parameters depend on characteristics of
the source, in particular on the relative frequency of topic changes.
Since broadcast news programs come from identified sources it
is useful and legitimate to optimize these parameters for 
each show type.\footnote{Shows in the 1997 BN corpus come from
eight sources: ABC World News Tonight, CNN Headline News,
CNN Early Prime, PRI The World, CNN Prime News, CNN The World Today,
C-SPAN Public Policy, and C-SPAN Washington Journal. Six of these
occurred in the test set.}
We therefore optimized the global parameter for each model to minimize
the segmentation cost on the training corpus 
(after training all other model parameters in a source-independent fashion).

Compared to a baseline using source-independent global TSP and threshold,
the source-dependent models showed between 5 and 10\% relative error 
reduction.  All results reported below use the
source-dependent approach.

\subsection{Segmentation Results}
	\label{sec:results}

Table~\ref{primaryresults} shows the results for both individual
knowledge sources (words and prosody), as well as for the combined 
models (decision tree and HMM).
It is worth noting
that the prosody-only results were obtained by running the combined HMM
without language model likelihoods; this approach gave better performance
than using the prosodic decision trees directly as classifiers.

Both word- and time-based metrics are 
given; they exhibit generally very similar results.
Another dimension of the evaluation is the use of correct word 
transcripts (forced alignments) versus automatically recognized words.
Again, results along this dimension are very similar, with some exceptions
noted below.

\begin{table}[t]
\tcaption{Summary of error rates with the language model only (LM),
the prosody model only
(PM), the combined decision tree (CM-DT), and the combined HMM (CM-HMM).
(a) shows word-based error metrics, (b) shows time-based error metrics.
In both cases a ``chance'' classifier that labels all potential boundaries
as nontopic would achieve 0.3 weighted segmentation cost.}
\label{primaryresults}
\begin{itemize}
\item[(a)]
\begin{tabular}[t]{lcccccc}
& \multicolumn{3}{c}{Error Rates on Forced Alignments}
& \multicolumn{3}{c}{Error Rates on Forced Alignments} \\
Model &$P_{\it Miss}$&$P_{\it FalseAlarm}$&$C_{\it Seg}$&$P_{\it Miss}$&$P_{\it FalseAlarm}$&$C_{\it Seg}$\\
\hline
Chance & 1.0\0\0\0 & 0.0\0\0\0 & 0.3\0\0\0 & 1.0\0\0\0 & 0.0\0\0\0 & 0.3\0\0\0 \\
LM &0.4847&0.0630&0.1895&0.4978&0.0577&0.1897\\
PM &0.4130&0.0596&0.1657&0.4125&0.0705&0.1731\\
CM-DT &0.4677&0.0260&0.1585&0.4891&0.0146&0.1569\\
CM-HMM &0.3339 &0.0536&0.1377&0.3748&0.0450&0.1438\\
\end{tabular}

\item[(b)]
\begin{tabular}[t]{lcccccc}
& \multicolumn{3}{c}{Error Rates on Forced Alignments}
& \multicolumn{3}{c}{Error Rates on Forced Alignments} \\
Model &$P_{\it Miss}$&$P_{\it FalseAlarm}$&$C_{\it Seg}$&$P_{\it Miss}$&$P_{\it FalseAlarm}$&$C_{\it Seg}$\\
\hline
Chance & 1.0\0\0\0 & 0.0\0\0\0 & 0.3\0\0\0 & 1.0\0\0\0 & 0.0\0\0\0 & 0.3\0\0\0 \\
LM &0.5260&0.0490&0.1921&0.5361&0.0415&0.1899\\
PM &0.3503&0.0892&0.1675&0.3846&0.0737&0.1669\\
CM-DT &0.5136&0.0210&0.1688&0.5426&0.0125&0.1715\\
CM-HMM &0.3426&0.0496&0.1375&0.3746&0.0475&0.1456\\
\end{tabular}
\end{itemize}
\end{table}

Comparing the individual knowledge sources, we observe that prosody alone
does somewhat better than the word-based HMM alone.
The types of errors made differ consistently:  the prosodic model has
a higher false alarm rate, while the word-LMs have more miss errors.
\cbstart
The prosodic model shows more false alarms because many regular sentence 
boundaries often show characteristics similar to those of topic boundaries.
It also suggests that both models could be combined by letting
the prosodic model selects candidate topic boundaries
that are then filtered using lexical information.
\cbend

The combined models generally improve on the individual knowledge sources.%
\footnote{\label{note:bad-dt}
The exception is the time-based evaluation of the combined decision
tree.  We found that the posterior probability threshold optimized on
the training set works poorly on the test set for this model architecture
and the time-based evaluation.
The threshold that is optimal on the {\em test} set achieves
$C_{\it seg} = 0.1651$.  Section~\ref{sec:dt-combined} gives a possible 
explanation for this result.}
In the word-based evaluation, the combined decision tree (DT) reduced 
overall segmentation cost by 19\% over the language model on true words
(17\% on recognized words).
The combined HMM gave even better results: 
27\% and 24\% improvement in the error rate over the
language model for the true and recognized words, respectively.

Looking again at the breakdown of errors, we can see that the two model
combination approaches work quite differently:  the combined DT
has about the same miss rate as the LM, but lower false alarms.
The combined HMM, by contrast, combines a miss rate as low as (or
lower than) that of the prosodic model with the lower false alarm rate of
the LM, suggesting that the functions of the two knowledge sources are
complementary,
\cbstart
as discussed above.
Furthermore, the different error patterns of the two combination approaches
suggest that further error reductions could be achieved by combining 
the two hybrid models.\footnote{
Such a combination of combined models was suggested by one of the 
reviewers; we hope to pursue it in future research.}
\cbend

\begin{figure}[t]
\psfig{figure=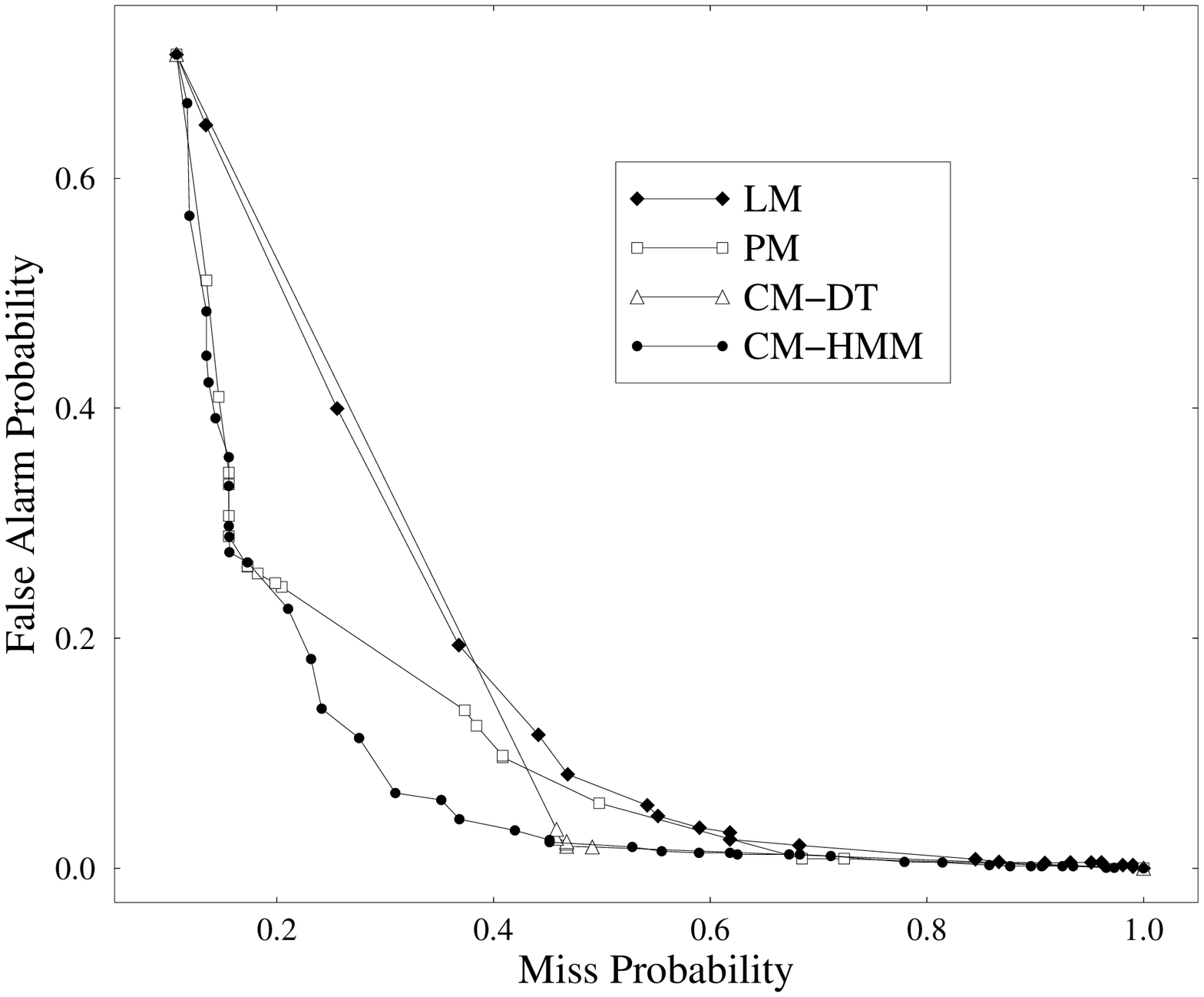,angle=270,width=0.9\textwidth}
\vskip 0.5\baselineskip
\caption{False alarm versus miss probabilities (word-based metrics)
for automatic topic segmentation from known words (forced alignments).
The segmenters used were a words-only HMM (LM), a prosody-only HMM (PM),
a combined decision tree (CM-DT), and a combined HMM (CM-HMM).}
\label{fig:roc}
\end{figure}

The trade-off between false alarms and miss probabilities is shown in
more detail in Figure~\ref{fig:roc}, which plots the two 
error metrics against each other.
Note that the false alarm rate does not reach 1 because the segmenter
is constrained by the chopping algorithm: the pause criterion prevents
the segmenter from hypothesizing topic boundaries everywhere.

\subsection{Decision Tree for the Prosody-only Model}
	\label{sec:dt-prosody-only}

\newsavebox{\rotbox}
\newlength{\oldtextwidth}
\setlength{\oldtextwidth}{\textwidth}
\begin{figure}[p]
\sbox{\rotbox}{%
\begin{minipage}{\textheight}
\psfig{figure=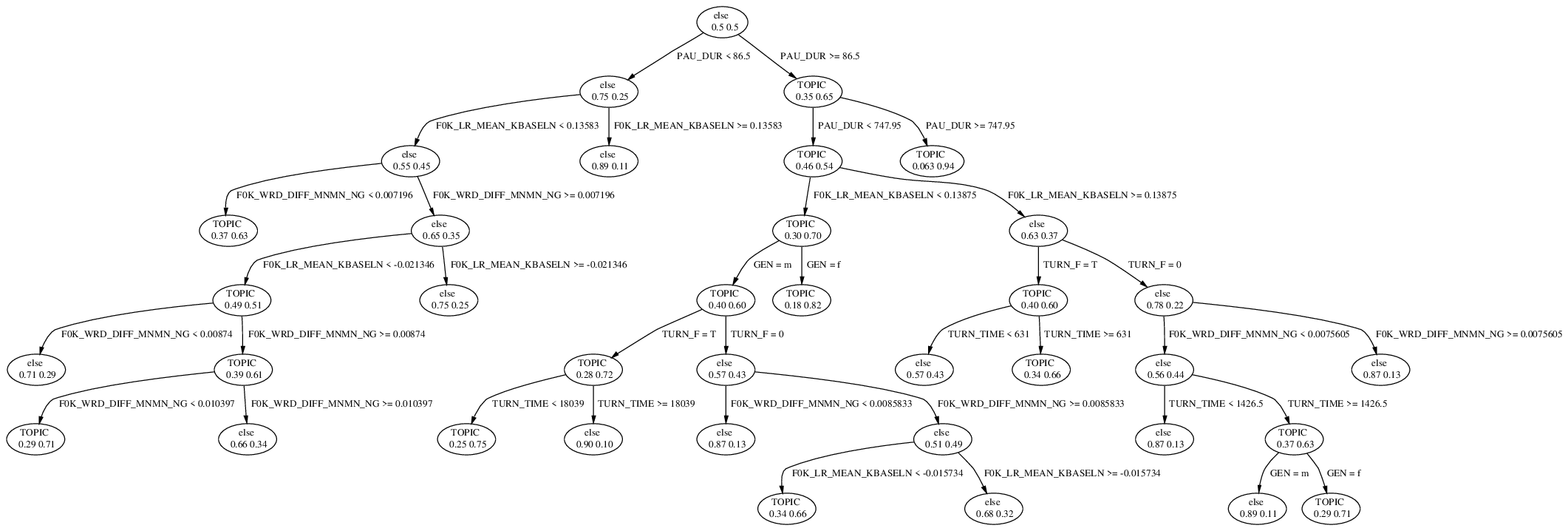,height=0.9\oldtextwidth,width=0.95\textheight,angle=270}
\vskip 0.5\baselineskip
\caption{The decision tree using prosodic features only.}
\label{fig:dt}
\end{minipage}}
\rotate[l]{\usebox{\rotbox}}
\end{figure}

Feature subset selection was run with an initial set of
73 potential features,
which the algorithm reduced to a set of 7 nonredundant features
helpful for the topic segmentation task.
The full decision tree learned is shown in Figure~\ref{fig:dt}.
We can identify four different kinds of features used in the tree, listed 
below.  For each feature type, we give the feature names found in the tree
and the {\bf relative feature usage},
an approximate measure of feature importance \cite{ShribergEtAl:eurospeech97}.
Relative feature usage is computed as the relative frequency with which
features of a given type are queried in the tree, over a held-out test set.

% The new features usages:
%     106dh-52 PAU_DUR              0.42649 
%     106dh-52 F0K_LR_MEAN_KBASELN  0.25380 
%     106dh-52 F0K_WRD_DIFF_MNMN_NG 0.10538 
%     106dh-52 TURN_F               0.09339 
%     106dh-52 GEN                  0.06826 
%     106dh-52 TURN_TIME            0.05267 

\begin{enumerate}
\item {\bf Pause duration} ({\tt PAU\_DUR}, 42.7\% usage).
This feature is the duration of the
nonspeech interval occurring at the boundary. The importance
of pause duration is underestimated here because,
as explained earlier, pause durations are already used
during the chopping process, so that the decision tree is applied only
to boundaries exceeding a certain duration.  Separate experiments
using boundaries below our chopping threshold show that the tree also 
distinguishes shorter pause durations for segmentation decisions.

\item {\bf F0 differences across the boundary}
({\tt F0K\_LR\_MEAN\_KBASELN} and {\tt F0K\_WRD\_DIFF\_MNMN\_NG}, 35.9\% usage).
These features compare the mean F0 of the word preceding the boundary
(measured from voiced regions within that word) to either the
speaker's estimated baseline F0 ({\tt F0K\_LR\_MEAN\_KBASELN}) or to the
mean F0 of the word following the boundary ({\tt F0K\_WRD\_DIFF\_MNMN\_N}).
Both features were computed based on a log-normal scaling of F0.
Other measures (such as minimum or maximum F0 in the word or preceding
window) as well as other normalizations (based on F0
toplines, or non-log-based scalings) were included in the initial feature set,
but were not selected in the
best-performing tree. The baseline feature captures a pitch range
effect, and is useful at boundaries where the speaker changes (since
range here is compared only within-speaker). The second feature
captures the relative size of the pitch change at the boundary, but of
course is not meaningful at speaker boundaries.

\item {\bf Turn features} ({\tt TURN\_F} and {\tt TURN\_TIME}, 14.6\% usage).
These features reflect the change of speakers.
{\tt TURN\_F} indicates whether a speaker change occurred at the
boundary, while {\tt TURN\_TIME} measures the time passed since the
start of the current turn.

\item {\bf Gender} ({\tt GEN}, 6.8\% usage).
This feature indicates the speaker gender right before a potential boundary.

\end{enumerate}

Inspection of the tree reveals that the purely prosodic features
(pause duration and F0 differences) are used as the prosody literature
suggests.
The longer the observed pause, the more likely a boundary corresponds
to a topic change.
Also, the closer a speaker comes to his or her F0 baseline, or the larger
the difference to the F0 following a boundary, the more likely a topic
change occurs.  These features thus correspond to the well-known
phenomena of boundary tones and pitch reset that are generally associated with 
sentence boundaries \cite{Vaissiere:83}. We found these indicators of 
sentences boundaries to be particularly pronounced at topic boundaries.

While turn and gender features are not prosodic features per se,
they do interact closely with them since prosodic measurements must
be informed by and carefully normalized for speaker identity and gender,%
\footnote{For example, the features that measure F0 differences across
boundaries do not make sense if the speaker changes at the boundary.
Accordingly, we made such features undefined for the decision tree
at turn boundaries.}
and it is therefore natural to include them in a prosodic classifier.
Not surprisingly, we find that turn boundaries are positively correlated
with topic boundaries, and that topic changes become more likely the 
longer a turn has been going on.

Interestingly, speaker gender is used by the decision tree for several
reasons.
One reason is stylistic differences
between males and females in the use of F0 at topic boundaries.  This
is true even after proper normalization, e.g., equating the
gender-specific nontopic boundary distributions.  In addition, we
found that nontopic pauses (i.e., chopping boundaries) are more
likely to occur in male speech.  It could be
that male speakers in BN are assigned longer topic segments on average,
or that male speakers are more prone to pausing in general, or that
male speakers dominate the spontaneous speech portions, where pausing is
naturally more frequent.  The details of this gender effect await 
further study.

\subsection{Decision Tree for the Combined Model}
	\label{sec:dt-combined}

\begin{figure}[t]
\psfig{figure=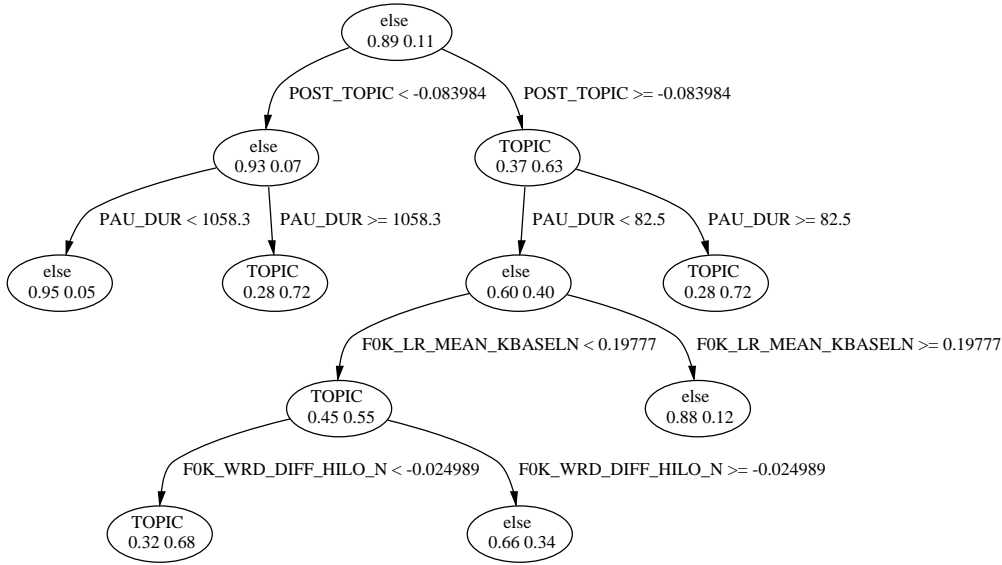,angle=270,width=\textwidth}
\vskip 0.5\baselineskip
\caption{The decision tree of the combination model.}
\label{fig:dt2}
\end{figure}

Figure~\ref{fig:dt2} depicts the decision tree that combines the
HMM language model topic decisions with prosodic features
(see Section~\ref{combined-dt}).
Again, we list the features used  with their relative feature usages.

% The feature usages are:
%        105dh PAU_DUR              0.49272 
%        105dh POST_TOPIC           0.49272 
%        105dh F0K_LR_MEAN_KBASELN  0.00885 
%        105dh F0K_WRD_DIFF_HILO_N  0.00572 

\begin{enumerate}
\item {\bf Language model posterior} ({\tt POST\_TOPIC}, 49.3\% usage).
This is the posterior probability $P(\B_i = {\bf yes}| W)$ computed from the 
HMM.

\item {\bf Pause duration} ({\tt PAU\_DUR}, 49.3\% usage). 
This feature is the same as described for the prosody-only model.

\item {\bf F0 differences across the boundary}
({\tt F0K\_WRD\_DIFF\_HILO\_N and F0K\_LR\_MEAN\_KBASELN}, 1.4\% usage). 
These features are similar to those found for the prosody-only tree.
The only difference is that for the first feature, the
comparison of F0 values across the boundary is done by taking the
maximum F0 of the previous word and the minimum F0 of the following
word---rather than the mean for both cases.

\end {enumerate}

The decision tree found for the combined task is smaller and uses fewer
features than the one trained with prosodic features only, for two reasons.
First, the LM posterior feature
is found to be highly informative, superseding the selection of many of 
the low-frequency features previously found.  Furthermore,
as explained in Section~\ref{combined-hmm}, the
prosody-only tree was trained on a downsampled dataset that equalizes
the priors for topic and nontopic boundaries, as required for 
integration into the HMM.
A welcome side effect of this procedure is that it forces the tree
to model the less frequent class (topic boundaries) in much greater
detail than if the tree were trained on the raw class distribution,
as is the case here.

Because of its small size, the tree in Figure~\ref{fig:dt2} is particularly
easy to interpret.  
The top-level split is based on the LM posterior.
The right branch handles cases where words are highly indicative of 
a topic boundary.  However, for short pauses the tree queries further
prosodic features to prevent false alarms.
Specifically, short pauses must be accompanied both by an F0 close to the
speaker's baseline and by a large F0 reset to be deemed topic boundaries.
Conversely, if the LM posteriors are low (left top-level branch),
but the pause is very long, the tree still outputs a topic boundary.

\subsection{Comparison of Model Combination Approaches}

Results indicate that the model combination approach
using an HMM as the top-level model works better than the 
combined decision tree.  While this result deserves more investigation
we can offer some preliminary insights.

We found it difficult to set the posterior probability thresholds for 
the combined decision tree in a robust way.  
As shown by the ``CM-DT'' curve in Figure~\ref{fig:roc},
there is a large jump in the false alarm/miss trade-off for the combined 
tree, in contrast to the combined HMM approach, which controls the trade-off
by a changing topic switch penalty.
This occurs because posterior probabilities from the decision
tree do not vary smoothly; rather, they vary in steps corresponding to
the leaves of the tree.  The discontinuous character of the thresholded
variable  makes it hard to estimate a threshold on the training data
that performs robustly on the test data.
This could account for the poor result on the time-based metrics
for the combined tree (where the threshold optimized on the training data
was far from optimal on the test set; see footnote~\ref{note:bad-dt}).
The same phenomenon is reflected in the fact that the prosody-only tree
gave better results when embedded in an HMM without LM likelihoods than 
when used by itself with a posterior threshold.

\cbstart
\subsection{Contributions of Different Feature Types}

We saw in Section~\ref{sec:dt-prosody-only} that pause duration is by far
the single most important feature in the prosodic decision tree.
Furthermore, speaker changes are queried almost as often as the 
F0-related features.  Pause durations can be obtained using 
standard speech recognizers, and are in fact used by many current TDT systems
(see Section~\ref{sec:comparison}).
Speaker changes are not prosodic features per se,
and would be detected independently from the prosodic features proper.
To determine if prosodic measurements beyond
pause and speaker information improve topic segmentation accuracy, we tested
systems that consisted of the HMM with the usual topic LMs, plus 
a decision tree that had access only to various subsets of pause- and
speaker-related features, without using any of the F0-based features.
Decision tree and HMM were combined as described in
Section~\ref{combined-hmm}.

\begin{table}[t]
\tcaption{Segmentation error rates with the language model only (LM),
  the combined HMM using all prosodic features (CM-HMM-all), the
  combined HMM using only pause duration and turn features
  (CM-HMM-pause-turn), and using only pause-duration, turn, and gender
  features (CM-HMM-pause-turn-gender).}
\label{pauseonly}
\begin{tabular}{lc}
Model & $C_{seg}$\\
\hline
LM &0.1895\\
CM-HMM-pause-turn&0.1519\\
CM-HMM-pause-turn-gender&0.1511\\
CM-HMM-all&0.1377\\
\end{tabular}
\end{table}

Table \ref{pauseonly} shows the results
of the system using only topic language models (LM) as well as
combined systems using all prosodic features (CM-HMM-all),
only pause duration and turn features (CM-HMM-pause-turn), and using only
pause duration, turn, and gender features (CM-HMM-pause-turn-gender).
These results show that by using only pause duration, turn, and gender
features, it is indeed possible to obtain
better results (20\% reduced segmentation cost)
than with the lexical model alone, with gender making only
a minor contribution.
However, we also see that a substantial further improvement (9\% relative)
is obtained by adding F0 features into the prosodic model.
\cbend

\subsection{Results Compared to Other Approaches}
	\label{sec:comparison}

Because our work focused on the use of prosodic information and 
required detailed linguistic annotations
(such as sentence punctuation, turn boundaries, and speaker labels),
we used data from the LDC 1997 BN corpus to form the training set
for the prosodic models and the (separate) test set used for evaluation.
This choice was crucial for the research, but unfortunately complicates
a quantitative comparison of our results to other TDT segmentation systems.
The recent TDT2 evaluation used a different set of 
broadcast news data that postdated the material used by
us, and was generated by a different speech recognizer
(although with a similar word error rate) \cite{TDT2:darpa99}.
Nevertheless we have attempted to calibrate our results with respect
to these TDT2 results.\footnote{
Since our study was conducted, a third round of TDT benchmarks (TDT3)
has taken place \cite{TDT3}.  However, for TDT3 the topic segmentation
evaluation metric was modified and the most recent results are thus not 
directly comparable with those from TDT2 or the present study.}
We have not tried to compare our
results to research outside the TDT evaluation framework.
In fact, other evaluation methodologies differ too much
to allow meaningful quantitative comparisons across publications.

We wanted to ensure that the TDT2 evaluation test set was 
comparable in segmentation difficulty to our test set drawn from the 1997 BN
corpus, and that the TDT2 metrics behaved similarly on both sets.
To this end, 
we ran an early version of our words-only segmenter on both test sets.
As shown in Table~\ref{tab:comparison}, not only are the results on
recognized words quite close, but the optimal false alarm/miss trade-off 
is similar as well, indicating that the two corpora have roughly similar
topic granularities.

\begin{table}[t]
\tcaption{Word-based segmentation error rates for different corpora.
Note that a hand-transcribed (forced alignment) version of the TDT2 test
set was not available.}
\label{tab:comparison}
\begin{tabular}{lcccccc}
& \multicolumn{3}{c}{Error Rates on Forced Alignments} &\multicolumn{3}{c}{Error Rates on Forced Alignments} \\
Test Set &$P_{\it Miss}$&$P_{\it FalseAlarm}$&$C_{\it Seg}$&$P_{\it Miss}$&$P_{\it FalseAlarm}$&$C_{\it Seg}$\\
\hline
TDT2 & NA&NA&NA&0.5509&0.0694&0.2139\\
BN'97 &0.4685&0.0817&0.1978&0.5128&0.0683&0.2017\\
\end{tabular}
\end{table}

% /home/spird41/stolcke/tdt/tdt2_dec98_official_results_19990109/segmentation/segmentation.htm
While the full prosodic component of our topic segmenter was not applied to
the TDT2 test corpus, we can compare the performance of a simplified 
version of SRI's segmenter to other evaluation systems
\cite{TDT2-NIST:darpa99}.
The two best-performing systems in the evaluation were those of
CMU \cite{Beeferman:99} with $C_{\it Seg} = 0.1463$, and 
Dragon \cite{Yamron:98,Dragon:darpa99} with $C_{\it Seg} = 0.1579$.
The SRI system achieved $C_{\it Seg} = 0.1895$.
All systems in the evaluation, including ours, used only information
from words and pause durations determined by a speech recognizer.

A good reference to calibrate our performance is the Dragon system,
from which we borrowed the lexical HMM segmentation framework.
Dragon made adjustments in its lexical modeling
that account for the improvements relative to the basic HMM structure
on which our system is based.
As described by \namecite{Dragon:darpa99},
a significant segmentation error reduction was obtained from optimizing the
number of topic clusters
(kept fixed at 100 in our system).
Second, Dragon introduced more supervision into the model training by
building separate LMs for segments that had been hand-labeled as
not related to news
(such as sports and commercials) in the TDT2 training corpus, which also
resulted in substantial improvements.
Finally, Dragon used some of the TDT2 training data for tuning
the model to the specifics of the TDT2 corpus.

In summary, the performance of our combined lexical-prosodic system with
$C_{\it Seg} = 0.1438$ is competitive with the best word-based systems
reported to date. More importantly, since we found the prosodic and
lexical knowledge sources to complement each other,
and since Dragon's improvements for TDT2 were confined to a better modeling of 
the lexical information,
we would expect that adding these improvements to our combined segmenter
would lead to a significant improvement in the state of the art.

\section{Discussion}
	\label{discussion}

Results so far indicate that prosodic information provides an
excellent source of information for automatic topic segmentation, both
by itself and in conjunction with lexical information.  Pause
duration, a simple prosodic feature that is readily available as a
by-product of speech recognition, proved highly effective in the
initial chopping phase, as well as being the most important feature
used by prosodic decision trees.  Additional, pitch-based prosodic
features are also effective as features in the decision tree.

The results obtained with recognized words (at 30\% word error rate)
did not differ greatly from those obtained with correct word
transcripts.  No significant degradation was found with the words-only
segmentation model, while the best combined  model exhibited about a 5\%
error increase with recognized words.  The lack of degradation on the
words-only model may be partly due to the fact that the recognizer generally 
outputs fewer words than contained in the correct transcripts,
biasing the segmenter toward a lower false alarm rate.
Still, part of the appeal of prosodic segmentation is that
it is inherently robust to recognition errors. This characteristic
makes it even more attractive for use in domains with higher error
rates due to poor acoustic conditions or more conversational speaking
styles.  It is especially encouraging that the prosody-only
segmenter achieved competitive performance.

It was fairly straightforward to modify the original Dragon HMM 
segmenter \cite{Yamron:98}, which is  based purely on topical word usage,
to incorporate discourse cues,
both lexical and prosodic. The addition of these discourse cues
proved highly effective, especially in the case of prosody.
The alternative knowledge source combination approach, using HMM posterior
probabilities as decision tree inputs, was also effective,
although less so than the HMM-based approach.  Note that the HMM-based
integration, as implemented here, makes more stringent assumptions about the 
independence of lexical and prosodic cues. The combined decision
tree, on the other hand, has some ability to model dependencies between
lexical and prosodic cues.  The fact that the HMM-based combination
approach gave the best results is thus 
indirect evidence that lexical and prosodic knowledge sources 
are indeed largely independent.

\begin{figure}[t]
\begin{itemize}
\item[(a)]
\fbox{\parbox[t]{0.8\textwidth}{\small
\setlength{\baselineskip}{\normalbaselineskip}
\ldots we have a severe thunderstorm watch two severe thunderstorm watches
and a tornado watch in effect the tornado watch in effect back here in
eastern colorado the two severe thunderstorm watches here indiana over
into ohio those obviously associated with this line which is already
been producing some hail i'll be back in a moment we'll take a look at
our forecast weather map see if we can cool it off in the east
will be very cold tonight minus seven degrees {\bf $<$TOPIC\_CHANGE$>$} \\

{\em LM probability: 0.018713}\\
{\em PM probability: 0.937276}\\
%ee970625.seg_647.367-413

karen just walked in was in the computer and found out for me that
national airport in washington d. c. did hit one hundred degrees today
it's a record high for them it's going to be uh hot again tomorrow but
it will begin to cool off the que question is what time of day is this
cold front going to move by your house if you want to know how warm
it's going to be tomorrow comes through early in the day won't be that
hot at all midday it'll still be into the nineties but not as hot as
it was today comes through late in the day you'll still be in the
upper nineties but some relief is on the way \ldots
}}
\item[(b)]
\fbox{\parbox[t]{0.8\textwidth}{\small
\setlength{\baselineskip}{\normalbaselineskip}
\ldots you know the if if the president has been unfaithful to his wife
and at this point you know i simply don't know any of the facts other
than the bits and pieces that we hear and they're simply allegations
at this point but being unfaithful to your wife isn't necessarily a
crime lying in an affidavit is a crime inducing someone to lie in an
affidavit is a crime but that occurred after this apparent taping so
i'll tell you there are going to be extremely thorny legal issues that
will have to be sorted out white house spokesman mike mccurry says the
administration will cooperate in starr's investigation 
{\bf $<$TOPIC\_CHANGE$>$} \\

{\em LM probability: 1.000000} \\
{\em PM probability: 0.134409}\\
%ed980121.seg_27.567-656

cubans have been waiting for this day for a long time after months of
planning and preparation pope john paul the second will make his first
visit to the island nation this afternoon it is the first pilgrimage
ever by a pope to cuba judy fortin joins us now from havana with
more \ldots
}}
\end{itemize}
\vskip 0.5\baselineskip
\caption{Examples of true topic boundaries where lexical and prosodic models
make opposite decisions.  (a) The prosodic model correctly predicts a topic
change, the LM does not.
(b) The LM predicts a topic change, the prosodic model does not.}
\label{fig:twoexamples}
\end{figure}

Apart from the question of probabilistic independence, it seems that
lexical and prosodic models are also complementary in the errors they
make.  This is manifested in the different distributions of
miss and false alarm errors discussed in Section~\ref{sec:results}.
It is also easy to find examples where the two models make complementary 
errors.  Figure~\ref{fig:twoexamples} shows two topic boundaries
that are missed by one model but not the other.

Several aspects of our model are preliminary or suboptimal in nature
and can be improved.
Even when testing on recognized words, we used 
parameters optimized on forced alignments.
This is suboptimal but convenient, since it avoids the need to
run word recognition on the relatively large training set.
Since results on recognized words are very similar to those on
true words we can conclude that not much was lost with this expedient.
Also, we have not yet optimized the chopping stage relative to the
combined model (only relative to the words-only segmenter).
The use of prosodic features other than pause duration for chopping should
further improve the overall performance.

The improvement obtained with source-dependent topic switch penalties and 
posterior thresholds suggests that more comprehensive source-dependent 
modeling would be beneficial.
In particular, both prosodic and lexical discourse cues are likely 
to be somewhat source specific (e.g., because of different show formats
and different speakers).  Given enough training data, it is straightforward 
to train source-dependent models.

\section{Conclusion}
	\label{conclusion}

We have presented a probabilistic approach for topic segmentation of
speech, combining both lexical and prosodic cues.
Topical word usage and lexical discourse cues are represented by
language models embedded in an HMM. Prosodic discourse cues,
such as pause durations and pitch resets, are modeled by a decision tree
based on automatically extracted acoustic features and alignments.
Lexical and prosodic features can be combined either in the HMM or 
in the decision tree framework.

Our topic segmentation model was evaluated on broadcast
news speech, and found to give 
competitive performance (around 14\% error according to the weighted TDT2
segmentation cost metric).
Notably, the segmentation accuracy of the prosodic model alone is
competitive with a word-based segmenter,
and a combined prosodic/lexical HMM achieves a substantial error
reduction over the individual knowledge sources.

\starttwocolumn

\small

\vspace{1em}

\begin{acknowledgments}
We thank Becky Bates, Madelaine Plauch{\'e}, Ze'ev Rivlin, Ananth Sankar,
and Kemal S\"onmez for invaluable assistance in preparing the data for this
study.
The paper was greatly improved as a result of comments by Andy Kehler,
Madelaine Plauch{\'e}, and the anonymous reviewers.
This research was supported by
DARPA and NSF under NSF grant IRI-9619921 and DARPA contract
no.~N66001-97-C-8544.  The views herein are those of the authors and
should not be interpreted as representing the policies of the funding
agencies.

\end{acknowledgments}

\small
\bibliography{all}

\end{document}